\documentclass[conference]{IEEEtran}

\usepackage{cite}
\usepackage{amsmath,amssymb,bbm,amsfonts}
\usepackage{algorithmic}
\usepackage{graphicx}
\usepackage{textcomp}
\usepackage{xcolor}
\usepackage{my_symbol}
\usepackage{booktabs}
\usepackage{multirow}
\usepackage[colorlinks=true,linkcolor=black,citecolor=blue,urlcolor=blue,backref=page]{hyperref}
\usepackage{comment}

\usepackage{pgfplots}
\pgfplotsset{compat=1.17}

\renewcommand*{\backrefalt}[4]{
	\ifcase #1 %
        {}
	\or
		cit. on p. #2.%
	\else
		cit. on pp. #2.%
	\fi}%

\newcommand\blfootnote[1]{%
  \begingroup
  \renewcommand\thefootnote{}\footnote{#1}%
  \addtocounter{footnote}{-1}%
  \endgroup
}

\usepackage{tikz}
\newcommand*\circled[1]{\tikz[baseline=(char.base)]{
            \node[shape=circle,draw,inner sep=2pt] (char) {#1};}}

\def\BibTeX{{\rm B\kern-.05em{\sc i\kern-.025em b}\kern-.08em
    T\kern-.1667em\lower.7ex\hbox{E}\kern-.125emX}}
\begin{document}

\title{Improved Image Classification with Manifold Neural Networks\\
}
\author{Caio F. Deberaldini Netto*, Zhiyang Wang\textsuperscript{\textdagger} and Luana Ruiz*}



\maketitle

\begin{abstract}
Graph Neural Networks (GNNs) have gained popularity in various learning tasks, with successful applications in fields like molecular biology, transportation systems, and electrical grids. These fields naturally use graph data, benefiting from GNNs' message-passing framework. However, the potential of GNNs in more general data representations, especially in the image domain, remains underexplored. Leveraging the manifold hypothesis, which posits that high-dimensional data lies in a low-dimensional manifold, we explore GNNs' potential in this context. We construct an image manifold using variational autoencoders, then sample the manifold to generate graphs where each node is an image. This approach reduces data dimensionality while preserving geometric information. We then train a GNN to predict node labels corresponding to the image labels in the classification task, and leverage convergence of GNNs to manifold neural networks to analyze GNN generalization. Experiments on MNIST and CIFAR10 datasets demonstrate that GNNs generalize effectively to unseen graphs, achieving competitive accuracy in classification tasks.
\end{abstract}


\section{Introduction}
\label{sec:intro}

\blfootnote{* Department of Applied Mathematics and Statistics, Mathematical Institute for Data Science (MINDS), Data Science and Artificial Intelligence Institute (DSAI), Johns Hopkins University, Baltimore, USA. E-mail: \{cnetto1, lrubini1\}@jh.edu}
\blfootnote{\textsuperscript{\textdagger} Department of Electrical and Systems Engineering, University of Pennsylvania, Philadelphia, USA. E-mail: zhiyangw@seas.upenn.edu}
The manifold hypothesis posits that high-dimensional data such as images lie on or near a low-dimensional manifold embedded within a high-dimensional ambient space. This assumption is widely used in machine learning to explain why certain algorithms can generalize well despite the high dimensionality of the input data \cite{bengio2013representation}. In the machine learning community, for instance, dimensionality reduction and manifold learning are research fields where the manifold hypothesis is applied with great success to reconstruct the low-dimensional geometrical structure of (sub-)manifolds from data \cite{belking2003laplacian_emap,mcinnes2018umap-software,watanabe2023spectralmap}.

Despite the success of these approaches, and the rise of geometric deep learning techniques such as graph and group-invariant neural networks \cite{cohen2016group,maron2018invariant,anderson2019cormorant,villar2021scalars,geiger2022e3nn,bronstein2017geometric}, the manifold structure underlying data without explicit geometry remains underexplored in deep learning. Inspired by recently introduced manifold neural networks (MNNs) \cite{wang2024gen_gnn,wang2024stability} and convergence results \cite{wang2024geometric} demonstrating that GNNs on geometric graphs sampled from them converge to MNNs, we propose a novel framework for image classification using GNNs.

Our first contribution is a method to build the manifold from image data. We do so by leveraging variational autoencoders (VAEs) \cite{kingma2014auto} which, unlike deterministic autoencoders, can produce meaningful image representations along a smooth and structured embedding space. After learning VAE image embeddings in an unsupervised manner, the graph is constructed by computing Gaussian kernel distances between embeddings, which are used as edge weights.

Our second contribution is a machine learning pipeline wherein images are seen as discrete points from the image manifold connected through a geometric graph, and embeddings are seen as a signal on this graph. Given this signal, we then train a GNN to predict node labels corresponding to the image labels in the classification task.

We validate our framework theoretically by proving that, on geometric graphs sampled from a manifold, GNNs have bounded generalization gap and, further, that this bound decreases with the graph size. This result is also verified empirically via numerical experiments on the MNIST and CIFAR10 datasets. Our numerical results show that GNNs achieve better generalization than a multilayer perceptron (MLP) trained on individual VAE embeddings, and that our method outperforms another GNN-based method in which graphs are built by interpreting image pixels as graph nodes \cite{dwivedi2023benchmarking}.
\section{Background}
\label{sec:background}

Before diving into our main contribution and method,  
we introduce preliminary definitions relating to graphs, graph neural networks, and manifold neural networks.

\subsection{Graph Signals, Graph Convolutions, Graph Neural Networks}

A \textit{graph} $\ccalG = (\ccalV, \ccalE, \ccalW)$ is defined as a triplet composed by a set of nodes \ccalV, where $N = |\ccalV|$ is the number of nodes, a set of edges $\ccalE \subseteq \ccalV \times \ccalV$, where $(i, j) \in \ccalE$ if nodes $i$ and $j$ are connected, and a function $\ccalW\colon \ccalE \to \reals$ that attributes weights to edges. 

\noindent \textbf{Graph signals.} Here, graphs are endowed with signals. More precisely, graph signals are defined as vectors $\bbx \in \reals^N$, where $x_i$ corresponds to the value of the signal at node $i$.

Given an undirected graph $\ccalG$, the \textit{graph shift operator} (GSO) $\bbS \in \reals^{N \times N}$ is defined as a symmetric matrix s.t. $\bbS_{i,j} \neq 0$ for every $(i, j) \in \ccalE$, and $\bbS_{i, j} = 0$ otherwise. The GSO operates on graph signals as $\bbS\bbx$ and, intuitively, it propagates/diffuses the graph signal through the nodes by aggregating the information of each node's neighborhood. Common examples of GSOs are the graph adjacency $\bbA$,
$$
\begin{cases}
    \bbA_{i, j} = 1, \mbox{ if } (i, j) \in \ccalE \\
    \bbA_{i, j} = 0 \mbox{ otherwise,}
\end{cases}
$$ 
the graph Laplacian $\bbL=\bbD-\bbA$, with $\bbD_{i, j} = \bbA_{i, :}\mathbbm{1}_{N}, i = j$, and $\bbD_{i, j} = 0$ otherwise, and their normalized versions \cite{ortega2018graph}. In this work, the chosen graph shift operator is the graph Laplacian, $\bbS = \bbL$. 

\noindent \textbf{Graph convolutional filters.} Given a graph signal $\bbx$ and a GSO $\bbS$, the graph convolutional filter $\bbh: \reals^N \to \reals^N$ is defined as
\begin{equation} \label{eqn:graph_convolution}
    \bbh(\bbS)\bbx = \sum_{k=0}^{K-1}h_{k}\bbS^{k}\bbx,
\end{equation}
i.e., it is a polynomial function of the GSO parameterized by  
coefficients $\{h_k\}_{k=0}^{K-1}$ and operating on graph signals \cite{segarra17-linear, ruiz2021gnns_stab_transf}.

\noindent \textbf{Graph neural networks.} One can define a \textit{graph neural network} as a stack of layers, each consisting of graph convolutional filters followed by point-wise non-linear transformations $\sigma\colon \reals \to \reals$. Precisely, we define the $l$th layer of a GNN as
\begin{equation} \label{eqn:gnn_layer}
    \bbX^l = \sigma\left(\sum_{k=0}^{K-1}\bbS^{k}\bbX^{l-1}\bbH_{k}^{l}\right),
\end{equation}
where $\bbX^{l-1} \in \reals^{N \times d_{l-1}}$ is the layer input and $\bbX^l \in \reals^{N \times d_l}$ is the layer output with $d_{l-1}$ input features and $d_l$ output features respectively, and $\bbH_{k}^{l} \in \reals^{d_{l-1} \times d_{l}}$ the filter coefficient matrix of the $l$th layer, which is learned.

For succinctness, throughout this paper the notation used for a GNN will be that of a function $\Phi(\bbX; \ccalH, \bbS)$, where $\ccalH=\{\bbH_k^l\}_{l,k}$ is the set of graph filter coefficients at all layers.

\subsection{Manifold Signals, Manifold Convolutions, Manifold Neural Networks}

Let $\ccalM$ be an $m$-dimensional, compact, and smooth submanifold embedded in $\reals^D$ with an induced uniform measure. More formally, $\ccalM$ is an $m$-dimensional smooth submanifold of $\reals^D$ if and only if every point $u\in \ccalM$ has an open neighborhood $U \subset \reals^D$ that can be mapped to some open subset $\Omega \subset \reals^m$ via a smooth map \cite{robbin2022introduction}. This is sometimes called the intrinsic definition of the manifold $\ccalM$.

Submanifolds of Euclidean space are locally Euclidean, in the sense that, in the vicinity of any point $u \in \ccalM$, the manifold and associated signals admit an Euclidean approximation via the so-called tangent space. The tangent space of $\ccalM$ at a point $u \in \ccalM$ is the collection of tangent vectors at $u$. A vector $\bbv \in \reals^D$ is a tangent vector of $\ccalM$ at $u$ if there exists a smooth curve $\gamma$ such that $\gamma(0) = u$ and $\dot{\gamma}(0)=\bbv$. In other words, a tangent vector can be seen as the derivative of a curve $\gamma: \reals \to \ccalM$. The tangent space at point $u$, denoted $T_u\ccalM$, is then \cite{robbin2022introduction}
\begin{equation*}
    T_u\ccalM = \{\dot{\gamma}(0)\ |\ \mbox{smooth }\gamma: \reals \to \ccalM \mbox{ , } \gamma(0)=u\} \text{.}
\end{equation*}
The collection of all tangent spaces at all points of the manifold $\ccalM$ is denoted $T\ccalM$ and called the tangent bundle.

\noindent \textbf{Manifold signals.} A manifold signal can be defined as a function over \ccalM, i.e., $f\colon \ccalM \to \reals$. We restrict our attention to $L^2$ functions over the manifold, i.e., $f \in L^2(\ccalM)$. 

Given smooth $f \in L^2(\ccalM)$, the gradient $\nabla f \in T\ccalM$ is the vector field satisfying $\smash{\langle \nabla f(u), \bbv \rangle = \frac{\D}{\D t}|_{t=0} \gamma(t)}$ for any tangent vector $\bbv \in T_u \ccalM$ and any smooth curve $\gamma$ such that $\gamma(0)=u$ and $\dot{\gamma}(0)=\bbv$ \cite{petersen2006riemannian}. Conversely, given a smooth vector field $F \in T\ccalM$ and an orthonormal basis $\bbe_1, \ldots, \bbe_D$ of $T_u\ccalM$, we define the divergence $\Delta F \in \ccalC^\infty (\ccalM)$ as $\Delta F = \sum_{i=1}^D \langle \partial_i F, \bbe_i\rangle$.

The composition of the gradient and divergence operators yields the Laplace-Beltrami (LB) operator $\ccalL: L^2(\ccalM) \to L^2(\ccalM)$, defined as \cite{berard2006spectral}
\begin{equation}
\ccalL f = - \Delta \left( \nabla f \right) \text{.}
\end{equation}
This operator appears in mathematical models of various physical phenomena, including wave propagation, heat diffusion, and the movement of quantum particles. Here, we are particularly interested on its role in the heat equation, which allows defining a \textit{manifold shift operator} (MSO) $ e^{-\ccalL } f$ diffusing the signal information $f$ through the manifold $\ccalM$ analogously to the GSO \cite{wang2024stability}.

\noindent \textbf{Manifold convolutional filters.} Henceforth, we can define \textit{manifold convolutional filters} as follows
\begin{equation} \label{eqn:manifold_conv}
    g = \bbh(\ccalL)f = \sum_{k=0}^{K-1}h_{k}e^{-k\ccalL}f,
\end{equation}
which, similarly to graph filters, are polynomial functions of the MSO parameterized by coefficients $\{h_k\}_{k=0}^{K-1}$, and applied to the manifold signal $f$ \cite{wang2024manifold}.

\noindent \textbf{Manifold neural networks.} At last, a \textit{manifold neural network} (MNN) can be defined as a stack of layers each consisting of manifold convolutional filters followed by point-wise non-linear transformations. Formally, the $l$th layer of an MNN is given by
\begin{equation}
    f^{l}(x) = \sigma\left(\sum_{k=0}^{K-1}e^{-k\ccalL}f^{l-1}(x)\bbH_{k}^{l}\right),
    \label{eqn:mnn}
\end{equation}
where $f^{l-1}: \ccalM \to \reals^{d_{l-1}}$ is the layer input and $f^l: \ccalM \to \reals^{d_l}$ the layer output with $d_{l-1}$ input features and $d_l$ output features respectively, and $\bbH_{k}^{l} \in \reals^{d_{l-1} \times d_l}$ are the learnable coefficients.
Similar to the notation used for GNNs, in the following we write the MNN as a function $\Phi(f; \ccalH, \ccalL)$, where $\ccalH$ groups the manifold filter coefficients at all layers.
\section{Exploiting Image Manifolds and Generalization of GNN\lowercase{s} via MNN\lowercase{s}}
\label{sec:method}

The manifold hypothesis posits that high-dimensional data lie on or near a low-dimensional manifold embedded within a high-dimensional ambient space ($m \ll D$). This assumption is widely used in machine learning to explain why certain algorithms can generalize well despite the high dimensionality of the input data. Specifically, we assume that that is the case for image data \cite{peyre2009manifold,osher2017low,brehmer2020flows,arbel2021generalized,miller2022graph}. Therefore, we need to define how we build and access that manifold, and how the incorporation of the manifold in the deep learning model affects its performance and generalization.

\subsection{Image manifolds}

Given a set of images $\{X_i\}_{i=1}^N$ sampled i.i.d. uniformly from an image space $\ccalX$, a natural approach to approximate their underlying manifold is to embed these images onto a lower dimensional space using machine learning techniques. 
A well-established architecture for learning data embeddings is the autoencoder, a customizable model consisting of an encoder and a decoder block \cite{wang2016auto}. The encoder $f_{\mbox{{\scriptsize enc}}}: \ccalX \to \reals^m$ reduces the data to a latent embedding $\bbz_i = f_{\mbox{{\scriptsize enc}}}(X_i)$ of specified size $m$, and the decoder $f_{\mbox{{\scriptsize dec}}}:\reals^m\to\ccalX$ takes this embedding and maps it back to original, ambient space, as $\tilde{X}_i = f_{\mbox{{\scriptsize dec}}}(\bbz_i)$. The functions $f_{\mbox{{\scriptsize enc}}},f_{\mbox{{\scriptsize dec}}}$ are learned by minimizing the distance between $\tilde{X}_i$ and $X_i$, 
\begin{equation} \label{eqn:sup_learning}
    \min_{f_{\mbox{{\scriptsize enc}}},f_{\mbox{{\scriptsize dec}}}} \sum_{i=1}^N \|f_{\mbox{{\scriptsize dec}}}(f_{\mbox{{\scriptsize enc}}}(X_i))-X_i\|_2^2 
\end{equation}

The encoder and decoder are typically deep networks tailored to the type of the data $X_i$ and the associated invariances---in the case of images, Convolutional Neural Networks (CNNs) for translation invariance or equivariance.  However, the data might also have invariances that are not known beforehand and hence not accounted for by the model used to parametrize the encoder and decoder. In these cases, autoencoders will often fail to map approximate invariants to close locations in embedding space, leading to poor approximations of the underlying manifold. 

Therefore, we propose to first learn that latent space by embedding the images from its original domain into the manifold using Variational Autoencoders (VAEs) \cite{kingma2014auto}.
VAEs differ from deterministic autoencoders in that instead of learning deterministic embeddings, they learn a Gaussian approximation $q(\bbz|X) = \ccalN(\bbz | \mu_z(X), \bbSigma_z(X))$ of the distribution $p(\bbz|X)$\footnote{For brevity, we omit details about the encoder and decoder parametrization and the loss function for the VAE, but a comprehensive introduction can be found in \cite{doersch2016tutorial}.}. Intuitively, this probabilistic framework contributes to a smoother embedding space, which is indeed observed empirically \cite{kusner2017grammar,dilokthanakul2016deep}. Further, the assumption of a Gaussian prior adds structure to the embedding space, and in practice it can be seen that the embedding dimensions are correlated with invariants of the data---provided the encoder and decoder are parametrized to preserve them---and other relevant features \cite{pu2016variational,dai2018syntax}.
Inspired by \cite{miller2022graph}, we train a CNNVAE, a VAE with a CNN as encoder/decoder, in an unsupervised way. 

\subsection{GNNs for image classification}
Given the learned embeddings $\bbz_i$, we can compute pairwise distances between them and construct a graph that can be processed by a GNN. Specifically, given a labeled dataset $\{\bbz_i, y_i\}_{i=1}^{N}$ of embedded images, where $y_i \in \{1,\ldots,C\}$ is the class label of image $i$, every sampled image is considered a node in the graph $\ccalG$. Given two images $i$ and $j$, the edge weight $\ccalW(i,j)$ is given by the Gaussian kernel distance
\begin{equation} \label{eqn:geom_graph}
    \ccalW(i,j) = \text{exp}\left(-\dfrac{||\bbz_i - \bbz_j||_{2}^2}{\sigma^2}\right),
\end{equation}
where $\sigma$ is the kernel width, which controls the neighborhood considered when propagating the nodes' information. 

We further define the $m$-dimensional graph signal $\bbZ \in \reals^{N \times m}$ supported on $\ccalG$, where the $i$th row of $\bbZ$ corresponds to the $i$th embedding $\bbz_i$, and the scalar graph signal $\bby \in \reals^N$, where the $i$th entry corresponds to the $i$th image label $y_i$. 
These signals are then used to learn a GNN $\Phi_\ccalH$ which approximates $\bby$ as $\tby = \Phi(\bbZ;\ccalH,\bbS)$ [cf. \eqref{eqn:gnn_layer}], where $\bbS$ is the Laplacian of graph \eqref{eqn:geom_graph}.

\subsection{Generalization of GNNs}
\label{subsec:gen_gnn}

Let $\ccalG$ be a graph with $N$ nodes sampled i.i.d. uniformly from the $m$-dimensional image manifold $\ccalM$, such that each node represents an image and, therefore, is endowed with a graph signal $\bbz_i \in \reals^m$, which is the image's embedding. In addition, consider $\bbZ \in \reals^{N\times m}$ to be the node feature matrix, i.e., the graph signal matrix supported on $\ccalG$, and $\bby \in \reals^N$ the column-vector of the classes that each node belongs to.

Therefore, if we suppose that the manifold hypothesis holds for that scenario, we can take advantage of the results that relate GNNs' output and MNNs' output trained on that manifold. Specifically, if we have a GNN $\Phi(\bbZ; \ccalH, \bbS)$ trained to predict each node/image class $\{y_i\}_{i=1}^N$, then Proposition 1 and Corollary 2 in \cite{wang2024gen_gnn} show that the GNN's output converges, respectively, in \textit{probability} and \textit{expectation} to the MNN's output 
$\Phi(f; \ccalH, \ccalL)$, under mild assumptions.

More formally, given a positive and Lipschitz continuous loss function, $\ell(\Phi(\bbZ; \ccalH, \bbS), \bby)$, the training of the GNN seeks to minimize the empirical risk defined as 
\begin{equation}
    P_{E}^* = \min_{\ccalH}R_E(\ccalH) = \ell(\Phi(\bbZ; \ccalH, \bbS), \bby).
\end{equation}

However, the GNN goal is to minimize the statistical risk
\begin{equation}
    P_{S}^* = \min_{\ccalH}R_S(\ccalH) = \mathbb{E}_{\bbZ \sim \mu^{N}}[\ell(\Phi(\bbZ; \ccalH, \bbS), \bby)].
\end{equation}

Then, as proposed by \cite{wang2024gen_gnn}, the generalization gap (GA), defined as $GA = P_{S}^* - P_{E}^*$, of the GNN can be bounded as follows

\begin{theorem}{\cite[Theorem~1]{wang2024gen_gnn}}
    Suppose we have a GNN trained on $(\bbZ, \ccalG)$ with $N$ nodes sampled i.i.d. uniformly over a $m$-dimensional $\mathcal{M}$, the generalization gap GA is bounded in probability at least $1 - \delta$ satisfying that,
    \[
    GA = \mathcal{O} \left( \left( \frac{\log \frac{N}{\delta}}{N} \right)^{\frac{1}{m+4}} + \left( \frac{\log N}{N} \right)^{\frac{1}{m+4}} \right).
    \]
    \label{thm:gnn_ga}
\end{theorem}


\noindent \textit{Proof}: See appendix.

\

This generalization gap depends on the size of the sampled graph $N$, i.e. the number of labeled images, as well as the underlying manifold dimension $m$. We observe that a GNN trained on a set of sampled images from the underlying image manifold can generalize to unseen graphs derived from the same image manifold. With these unseen graphs constructed from previously unlabeled image embeddings, the generalization capability demonstrates the GNN's ability to generalize and make predictions on new images.
\section{Experimental Results}
\label{sec:experiments}


\begin{table}[t]
    \centering
    \caption{Train and test accuracy on MNIST and CIFAR10 datasets. The subscript in our best model is the number of sampled nodes from the manifold to build the graph for each image.}
    \resizebox{\columnwidth}{!}{%
        \begin{tabular}{lcccc}
        \toprule
        \multirow{2}{*}{Model} & \multicolumn{2}{c}{MNIST} & \multicolumn{2}{c}{CIFAR10} \\
        \cmidrule(lr){2-3} \cmidrule(lr){4-5}
         & Test Acc. & Train Acc. & Test Acc. & Train Acc. \\
        \midrule
        GCN \cite{dwivedi2023benchmarking} & 90.12 $\pm$ 0.15 & 96.46 $\pm$ 1.02 & 54.14 $\pm$ 0.40 & 70.16 $\pm$ 3.43 \\
        $\text{GCN}_{[5]}$ (Ours) & 95.43 $\pm$ 0.11 & 95.93 $\pm$ 0.07 & 57.61 $\pm$ 0.19 & 86.26 $\pm$ 0.22 \\
        \bottomrule
        \end{tabular}
    }
    \label{tab:accuracy}
\end{table}

\begin{figure}[b]
    \centering
    \begin{tikzpicture}
        \begin{axis}[
            xlabel={Number of sampled nodes ($N$)},
            ylabel={$\text{Train}_\text{Acc.} - \text{Test}_\text{Acc.}$},
            xmin=0, xmax=55,
            ymin=-1, ymax=1.0,
            grid=major,
            xmode=linear,
            ymode=linear,
            legend pos=south west,
            ytick={-0.5, -0.2, 0.1, 1.0},
            xtick={5, 10, 20, 25, 50}
        ]
        \addplot[
            color=blue,
            mark=star,
            dashed,
            line width=1pt
        ] coordinates {
            (5, 0.49)
            (10, 0.07)
            (20, -0.17)
            (25, -0.31)
            (50, -0.43)
        };
        \addlegendentry{GCN}
        \addplot[
            color=red,
            mark=star,
            dashed,
            line width=1pt
        ] coordinates {
            (5, 0.91)
            (10, 0.91)
            (20, 0.91)
            (25, 0.91)
            (50, 0.91)
        };
        \addlegendentry{MLP}
        \end{axis}
    \end{tikzpicture}
    \caption{Accuracy difference between train and test set for an increasing number of sampled nodes for MNIST dataset. The generalization gap (GA) decreases as the number of nodes increases.}
    \label{fig:empirical_ga_mnist}
\end{figure}
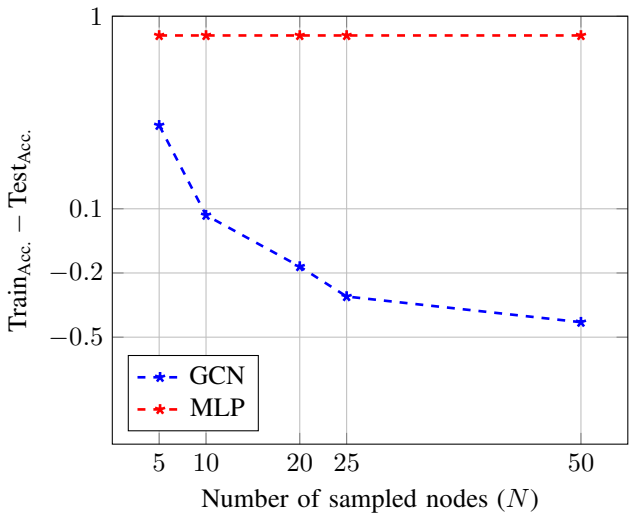

To assess the validity of our method, we show that the generalization bound presented in (\ref{subsec:gen_gnn}) holds for standard image classification benchmarks. Specifically, we use MNIST and CIFAR10 \cite{lecun1998gradient,krizhevsky2009learning}. The former is a dataset of images of handwritten digits in greyscale, comprising $60,000$ training samples and $10,000$ test samples. Our expectation with this dataset was that our models would have an almost perfect performance, and we could have a sanity check. On the other hand, the latter is a dataset of RGB images of 10 different objects, comprising $50,000$ training samples and $10,000$ test samples.

For both datasets, we first preprocess the data by training the CNNVAE. This first task is to encode the images into a latent space, and then reconstruct them. From experiments, the best latent space for MNIST has size $m = 128$, while for CIFAR10, $m = 4096$. 

After that, with the embedded images, we use a GNN to process the sampled manifold data and predict the class of the graph nodes. Precisely, for each embedded image we uniformly sample $N - 1$ images in the image's set (train or test), forming a graph with $N$ nodes. As previously stated, the graph signal or node feature matrix corresponds to the images' embeddings.

Empirical results are shown in Table \ref{tab:accuracy} for both datasets. We compare our model, a GCN \cite{kipf2017semi}, with the results obtained by the same GNN model but using the SLIC superpixel technique to build the graph structure from images presented in \cite{dwivedi2023benchmarking}. Since our focus so far is not to produce state-of-the-art results, we didn't fine-tune any hyperparameter for the models we implemented. On the contrary, we used the same principles used by the authors in \cite{dwivedi2023benchmarking} when defining our neural networks' hyperparameters. For instance, we used a small hidden size representation and a small number of layers, such that the model had between $100$k-$500$k parameters.

\begin{figure}[t]
    \centering
    \begin{tikzpicture}
        \begin{axis}[
            xlabel={Number of sampled nodes ($N$)},
            ylabel={$\text{Train}_\text{Acc.} - \text{Test}_\text{Acc.}$},
            xmin=0, xmax=55,
            ymin=15, ymax=35,
            grid=major,
            xmode=linear,
            ymode=linear,
            legend pos=south west,
            ytick={15, 20, 25, 35},
            xtick={5, 10, 20, 25, 50}
        ]
        \addplot[
            color=blue,
            mark=star,
            dashed,
            line width=1pt
        ] coordinates {
            (5, 28.65)
            (10, 26.60)
            (20, 23.74)
            (25, 22.69)
            (50, 18.96)
        };
        \addlegendentry{GCN};
        \addplot[
            color=red,
            mark=star,
            dashed,
            line width=1pt
        ] coordinates {
            (5, 33.86)
            (10, 33.86)
            (20, 33.86)
            (25, 33.86)
            (50, 33.86)
        };
        \addlegendentry{MLP}
        \end{axis}
    \end{tikzpicture}
    \caption{Accuracy difference between train and test set for an increasing number of sampled nodes for CIFAR10 dataset. The generalization gap (GA) decreases as the number of nodes increases.}
    \label{fig:empirical_ga_cifar10}
\end{figure}
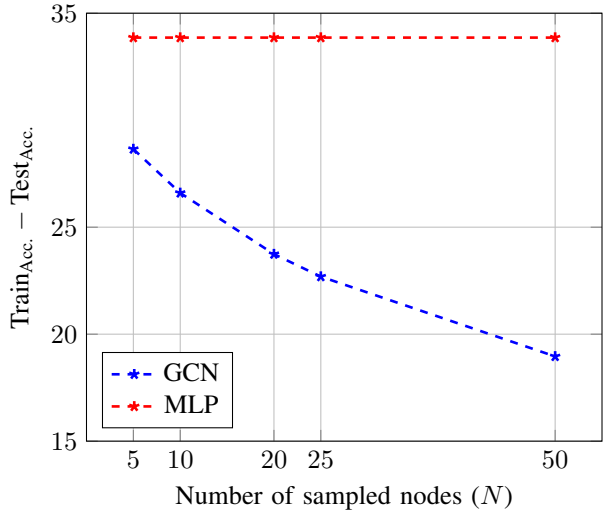

Our method had better results than that proposed in \cite{dwivedi2023benchmarking}, as seen in Table \ref{tab:accuracy}, and our GNN was able to reduce the gap between seen and unseen data, as seen in Figures \ref{fig:empirical_ga_mnist} and \ref{fig:empirical_ga_cifar10}, a result we expected seeing empirically, given the theoretical result showed in Theorem \ref{thm:gnn_ga}. 

Results were obtained after training 10 GCN models with 10 different seeds for 300 epochs. For the latter experiment we did the same for each number of sampled nodes in $\{5, 10, 20, 25, 50\}$, and we also showed the generalization gap for a vanilla MLP trained on the same set of embeddings.

\section{Conclusions}
\label{sec:conclusion}

In this work, we introduced a novel framework for image classification that exploits the manifold hypothesis by creating a geometric graph from image data embedded using a VAE. By treating these embeddings as graph signals, we applied GNNs to the classification tasks, achieving better generalization. Theoretical analysis confirmed that GNNs trained on graphs sampled from a manifold have a bounded generalization gap, decreasing as graph size grows. Experiments on MNIST and CIFAR10 showed our method's superiority over MLPs and pixel-based GNNs, opening new possibilities for deep learning with manifold representations, particularly in scenarios where the underlying geometry of data is not explicitly known.

\vfill\pagebreak

\bibliographystyle{IEEEbib}
\bibliography{myIEEEabrv,refs,bib_cumulative}

\vfill\pagebreak
\section*{Appendix}
\noindent\textit{A. Proof of Theorem 1}

\

To prove the generalization bound for the GNN, we first need to bring up two results that relate the output of a GNN $\Phi(\bbZ; \ccalH,\bbS)$ and an MNN $\Phi(f; \ccalH, \ccalL)$.

\

\begin{definition}
    A manifold signal $f$ is $\lambda_M$-bandlimited if for all eigenpairs $\{\lambda_i, \phi_i\}_{i=1}^\infty$ of the Laplace-Beltrami operator\footnote{The LB operator $\ccalL$ is self-adjoint and positive semidefinite (PSD). Hence, it admits a spectral decomposition.} $\ccalL$ when $\lambda_i > \lambda_M$, we have $\langle f, \phi_i\rangle = 0$.
\end{definition}

\

\begin{definition}
    A filter is a low-pass filter if its frequency response satisfies
    \begin{equation*}
        |\hhath(a)| = \ccalO(a^{-d}).
    \end{equation*}
\end{definition}

\begin{definition}
    A nonlinear activation function $\sigma(\cdot)$ is normalized Lipschitz-continuous if it satisfies
    \begin{equation*}
        |\sigma(x) - \sigma(y)| \leq |x - y|, \ \text{with} \ \sigma(0) = 0.
    \end{equation*}
\end{definition}

\begin{proposition}{\cite[Proposition~1]{wang2024gen_gnn}}\label{prop:gnn_mnn_conv}
    Let $\ccalM \subset \reals^D$ be an embedded manifold with Laplace-Beltrami operator $\ccalL$ and a $\lambda_M$-bandlimited manifold signal $f$. Consider a pair of graph and graph signal $(\ccalG, \bbZ)$ with $N$ nodes sampled i.i.d. uniformly over $\ccalM$. The graph Laplacian $\bbL$ is calculated with (\ref{eqn:geom_graph}). Let $\Phi(\cdot; \ccalH, \ccalL)$ be a single layer MNN on $\ccalM$ (\ref{eqn:mnn}) with single input and output features. Let $\Phi(\cdot; \ccalH, \bbL)$ be the GNN with the same architecture applied to the graph $\ccalG$. Then, with the filters as low-pass and nonlinearities as normalized Lipschitz continuous, it holds in probability at least $1 - \delta$ that
    \begin{multline*}
    \|\Phi(\bbP_{N} f; \ccalH, \bbL) - \bbP_{N}\Phi(f; \ccalH, \ccalL)\|_2 \leq \\ 
    C_1 \left( \frac{\log \frac{C_1 N}{\delta}}{N} \right)^{\frac{1}{m+4}} + C_2 \left( \frac{\log \frac{C_2 N}{\delta}}{N} \right)^{\frac{1}{m+4}} \theta_M^{-1} \\
    + C_3 \sqrt{\frac{\log(1/\delta)}{N}} + C_4 M^{-1},
    \end{multline*}
    where $\bbP_N \colon L^2(\ccalM) \to L^2(\ccalV(\ccalG))$ is an uniform sampling operator, $C_1, C_2, C_3$ and $C_4$ are constants and $\theta_M = \min_{i=1,2,...,M} |\lambda_i - \lambda_{i+1}|$.
\end{proposition}

\

\begin{corollary}{\cite[Corrolary~2]{wang2024gen_gnn}}\label{cor:gnn_mnn_expc_conv}
    The above difference bound between GNN and MNN also holds in expectation, since each node in $\ccalG$ is sampled i.i.d. uniformly over $\ccalM$
    \begin{multline*}
        \mbE\bigl[\|\Phi(\bbP_{N} f; \ccalH, \bbL) - \bbP_{N}\Phi(f; \ccalH, \ccalL)\|_2\bigl] \leq \\
        C'N^{\tfrac{1}{m+4}} + C''N^{-1/2} + C'''\left(\tfrac{\log N}{N}\right)^{\tfrac{1}{m+4}} + \bbarM e^{-N/C}\sqrt{N},
    \end{multline*}
    where $C', C''$ and $C'''$ are constants, and $\bbarM = 2\|\bbP_N f\|$.
\end{corollary}

\

Considering those results, suppose $\bbH_E \in \text{arg} \min_{\ccalH} R_E(\ccalH)$. Then, we have

\begin{multline}
    GA \leq R_S(\bbH_E) - R_E(\bbH_E) \\
    = \mathbb{E}_{\bbZ \sim \mu^{N}}[\ell(\Phi(\bbZ; \bbH_E, \bbL), \bby)] - \ell(\Phi(\bbZ; \bbH_E, \bbL), \bby).
\end{multline}

Adding and subtracting the term $\ell(\Phi(f; \bbH_E, \ccalL), g)$ we have the following

\begin{multline}
    GA \leq (\mathbb{E}_{\bbZ \sim \mu^{N}}[\ell(\Phi(\bbZ; \bbH_E, \bbL), \bby)] - \ell(\Phi(f; \bbH_E, \ccalL), g)) \\ 
    +  (\ell(\Phi(f; \bbH_E, \ccalL), g) - \ell(\Phi(\bbZ; \bbH_E, \bbL), \bby)).
\end{multline}

Taking the absolute value of the above inequality and applying the triangle inequality, we have

\begin{multline}\label{eqn:ga_1}
    GA \leq \underbrace{\biggl|\mbE_{\bbZ \sim \mu^{N}}\bigl[\ell(\Phi(\bbZ; \bbH_E, \bbL), \bby)\bigl] - \ell(\Phi(f; \bbH_E, \ccalL), g)\biggl|}_{\circled{1}} \\ 
    +  \underbrace{\biggl|\ell(\Phi(f; \bbH_E, \ccalL), g) - \ell(\Phi(\bbZ; \bbH_E, \bbL), \bby)\biggl|}_{\circled{2}}.
\end{multline}

Here, the loss function is assumed to be the $L_2$ loss. Therefore, the term \circled{1} in (\ref{eqn:ga_1}) can be written as

\begin{multline}\label{eqn:ga_2}
    \biggl|\mathbb{E}_{\bbZ}\bigl[\ell(\Phi(\bbZ; \bbH_E, \bbL), \bby)\bigl] - \ell(\Phi(f; \bbH_E, \ccalL), g)\biggl| \\
    = \biggl|\mathbb{E}_{\bbZ}\bigl[\|\Phi(\bbZ; \bbH_E, \bbL) - \bby\|\bigl] - \|\Phi(f; \bbH_E, \ccalL) - g\|_{\ccalM}\biggl|
\end{multline}

Now, by subtracting and adding the term $\mathbb{E}_{\bbZ}[\bbP_N\Phi(f; \bbH_E, \ccalL)]$ inside the expectation above, and remembering the fact that $\bby = \bbP_Ng$, the expectation term from (\ref{eqn:ga_2}) becomes the following

\begin{multline}\label{eqn:ga_3}
    \biggl|\mathbb{E}_{\bbZ}\bigl[\|\Phi(\bbZ; \bbH_E, \bbL) - \bby\|\bigl] - \|\Phi(f; \bbH_E, \ccalL) - g\|_{\ccalM}\biggl| \\
    \leq \biggl|\mathbb{E}_{\bbZ}\bigl[\|\Phi(\bbZ; \bbH_E, \bbL) - \bbP_N\Phi(f; \bbH_E, \ccalL)\|\bigl] + \\ 
    \mathbb{E}_{\bbZ}\bigl[\|\bbP_N\Phi(f; \bbH_E, \ccalL) - \bbP_Ng\|\bigl] - \|\Phi(f; \bbH_E, \ccalL) - g\|_{\ccalM}\biggl| \\
    \\
    \leq \biggl|\mathbb{E}_{\bbZ}\bigl[\|\Phi(\bbZ; \bbH_E, \bbL) - \bbP_N\Phi(f; \bbH_E, \ccalL)\|\bigl]\biggl| + \\
    \biggl|\mathbb{E}_{\bbZ}\bigl[\|\bbP_N\Phi(f; \bbH_E, \ccalL) - \bbP_Ng\|\bigl] - \|\Phi(f; \bbH_E, \ccalL) - g\|_{\ccalM}\biggl|
\end{multline}

The first term of equation (\ref{eqn:ga_3}) is bounded above using Corollary \ref{cor:gnn_mnn_expc_conv}. For the second term, we need to use a derivation of Theorem 19 in \cite{von2008consistency}. More specifically, given that the nodes in $\ccalG$ were sampled i.i.d. from $\ccalM$, then

\begin{equation}
    |\langle \bbP_Nf, \phi_i\rangle - \langle f, \phi_i\rangle| = \ccalO\left(\sqrt{\dfrac{\log 1/\delta}{N}}\right),
\end{equation}
for $\langle \cdot,\cdot \rangle$ being the inner product in $L^2$. This implies that $\bigl|\|\bbP_Nf\|^2 - \|f\|_{\ccalM}^2\bigl| = \ccalO\left(\sqrt{\dfrac{\log 1/\delta}{N}}\right)$, which indicates that  $\|\bbP_Nf\| = \|f\|_{\ccalM} + \ccalO\left(\dfrac{\log 1/\delta}{N}\right)^{\tfrac{1}{4}}$. Therefore, we have that

\begin{multline}\label{eqn:entr_bound}
    \mbP \biggl(\bigl|\|\bbP_N\Phi(f; \bbH_E, \ccalL) - \bbP_Ng\| - \|\Phi(f; \bbH_E, \ccalL) - g\|_{\ccalM}\bigl| \\
    \leq \ccalO\left(\dfrac{\log 1/\delta}{N}\right)^{\tfrac{1}{4}} \biggl) \geq 1 - \delta.
\end{multline}
An expectation value can also be devised based on the probability bound, similarly to the result in Corollary \ref{cor:gnn_mnn_expc_conv}, and then we can bound the second term of (\ref{eqn:ga_3}) as

\begin{multline}
    \mbE\bigl[\|\bbP_N\Phi(f; \bbH_E, \ccalL) - \bbP_Ng\| - \|\Phi(f; \bbH_E, \ccalL) - g\|_{\ccalM}\bigl] \\
    \leq CN^{-\tfrac{1}{4}} + \ccalO(e^{-N/C}\sqrt{N}).
\end{multline}

Now, assuming, again, that the loss function is the $L_2$ loss, we can rewrite the term \circled{2} in (\ref{eqn:ga_1}) as

\begin{multline}
    \bigl|\ell(\Phi(f; \bbH_E, \ccalL), g) - \ell(\Phi(\bbZ; \bbH_E, \bbL), \bby)\bigl| \\
    = \bigl|\|(\Phi(f; \bbH_E, \ccalL) - g)\| - \|(\Phi(\bbZ; \bbH_E, \bbL) - \bby)\|\bigl|.
\end{multline}

Adding and subtracting an intermediate term $\bbP_N\Phi(f; \bbH_E, \ccalL)$, and applying the triangle inequality we have the following with probability at least $1 - \delta$

\begin{multline}\label{eqn:ga_4}
    \bigl|\|\Phi(\bbZ; \bbH_E, \bbL) - \bbP_Ng\| - \|\Phi(f; \bbH_E, \ccalL) - g\|\bigl| \\
    = \bigl|\|\Phi(\bbZ; \bbH_E, \bbL) - \bbP_N\Phi(f; \bbH_E, \ccalL)\| +  \\
    \|\bbP_N\Phi(f; \bbH_E, \ccalL) - \bbP_Ng)\|\bigl| - \|(\Phi(f; \bbH_E, \ccalL) - g)\| \bigl| \\
    \\
    = \bigl|\|\Phi(\bbZ; \bbH_E, \bbL) - \bbP_N\Phi(f; \bbH_E, \ccalL)\| \bigl| + \\
    \bigl|\|\bbP_N\Phi(f; \bbH_E, \ccalL) - \bbP_Ng)\|\bigl| - \|(\Phi(f; \bbH_E, \ccalL) - g)\| \bigl|.
\end{multline}
The first term in (\ref{eqn:ga_4}) is bounded by Proposition \ref{prop:gnn_mnn_conv}, while the second term is bounded by (\ref{eqn:entr_bound}). Taking the leading orders from those bounds, we conclude that

\begin{equation}
    GA = \mathcal{O} \left( \left( \frac{\log \frac{N}{\delta}}{N} \right)^{\frac{1}{m+4}} + \left( \frac{\log N}{N} \right)^{\frac{1}{m+4}} \right).
\end{equation}

\hspace*{\fill}$\square$

\end{document}